\documentclass[sigconf,nonacm]{acmart}

\usepackage{fontawesome5}
\usepackage{xcolor}
\usepackage{makecell}
\usepackage{tabularx}
\usepackage{array}

\begin{document}

\title{WattCouncil: Context-Aware Household Energy Scenario Generation With Governed LLMs}

\author{Mohannad Takrouri}\authornote{Both authors contributed equally to this research.}
\orcid{0000-0001-8523-3601}
\affiliation{
\department{Machine Learning Department}\institution{Mohamed bin Zayed University of Artificial Intelligence}
\city{Abu Dhabi}
\country{United Arab Emirates}}
\email{mohannad.takrouri@mbzuai.ac.ae}

\author{Nicolas Cuadrado}\authornotemark[1]
\orcid{0000-0002-7791-2388}
\affiliation{
\department{Machine Learning Department}\institution{Mohamed bin Zayed University of Artificial Intelligence}
\city{Abu Dhabi}
\country{United Arab Emirates}}
\email{nicolas.avila@mbzuai.ac.ae}

\author{Martin Takac}
\orcid{0000-0001-7455-2025}
\affiliation{
\department{Machine Learning Department}\institution{Mohamed bin Zayed University of Artificial Intelligence}
\city{Abu Dhabi}
\country{United Arab Emirates}}
\email{martin.takac@mbzuai.ac.ae}

\renewcommand{\shortauthors}{Takrouri et al.}

\begin{abstract}
The accelerating shift toward low-carbon power systems, together with the widespread adoption of behind-the-meter technologies such as rooftop solar and electric vehicles, is placing new operational and analytical demands on electricity grids. At the same time, smart-grid research increasingly relies on machine learning (ML), yet progress is constrained by limited access to high-resolution household energy data due to privacy concerns, regulatory barriers, and collection costs. This work presents \textit{WattCouncil}, a data-generation framework in which household electricity demand is generated by a council of Large Language Model (LLM)–based agents operating in specialized roles to generate, audit, and validate structured energy scenarios under explicit cultural, temporal, and physical constraints. Rather than acting as static predictors, these agents serve as adaptive decision-makers within a governed pipeline. Motivated by studies highlighting the importance of contextual factors in energy use, our framework produces context-sensitive daily routines through a guided reasoning process that incorporates household composition, temporal factors, and environmental conditions. We evaluate the generated profiles against the detailed \href{https://www.scidb.cn/en/detail?dataSetId=311c824cbbf94f70b2e21a56f368bd5f}{CER} dataset, which contains over a year of load measurements for 4232 households together with survey-based socio-economic information. We further assess the consistency of the framework through ablation studies. \footnote{Source code is available at 
\url{https://github.com/Singularity-AI-Lab/wattcouncil}.
}
\end{abstract}


\maketitle
\section{Introduction} \label{sec:introduction}

Global energy consumption continues to grow alongside societal and economic development, intensifying efforts to improve energy efficiency and reduce carbon emissions to address climate change. At the same time, electricity systems are undergoing structural change driven by the widespread deployment of renewable energy sources (RES), whose inherent variability and intermittency introduce new forms of uncertainty for grid operation and analysis \cite{sinsel_challenges_2019, zhao_review_2025}. Traditional power grids were designed around centralized, dispatchable generation and relatively predictable consumption patterns. Modern energy systems, however, increasingly depend on a detailed understanding of how demand varies across time, context, and usage conditions, particularly as electrification and distributed resources reshape load dynamics. This evolving landscape has motivated energy research that prioritizes not only real-time operations but also the characterization and exploratory analysis of demand behavior as a foundation for planning, policy design, and comparative scenario assessment.

Although residential electricity consumption exhibits patterns that can be exploited by data-driven methods, many existing modeling approaches remain tied to limited historical data or rigid simulators that do not fully capture household-level variability. Empirical analyses have shown that electricity use in the residential sector is influenced by multiple determinants, including household demographics, dwelling characteristics, climate conditions, and appliance usage, and that these factors contribute to significant heterogeneity in consumption behavior both within and between households \cite{kostakis2020socio, ali2022demographics, guven2023analysing, lavin2025household}. For example, income, household size, and regional climate have been identified as major contributors to variation in total residential demand, while smart meter analyses reveal distinct daily load profiles associated with occupancy and dwelling characteristics \cite{rafiq2023analysis, grandjean2012review, jabbarReshapingSmartEnergy2021}. Similarly, research on residential load patterns has shown significant correlations among these factors, underscoring the need to incorporate them into accurate energy consumption models \cite{weiCharacterizingResidentialLoad2021}. Traditional simulation methods and bottom-up models often rely on assumptions that are difficult to scale or tailor to diverse contexts, and purely data-driven models require high-quality, well-annotated datasets. Gaps, missing values, and limited geographic coverage can reduce model generalization and lead to the omission of important behavioral structures. Together, these limitations underscore the need for methods that can generate controlled, contextually informed synthetic consumption data to support exploratory analysis and scenario reasoning beyond the scope of available empirical sources.

Large Language Models (LLMs) have emerged as powerful tools for synthesizing data and knowledge across diverse domains \cite{tanLargeLanguageModels2024}, including recent exploratory applications in the energy sector. After the introduction of LLM-based agents into the research community \cite{park2023generative}, new paradigms have been proposed for leveraging the implicit knowledge encoded in large models through structured interaction and coordination. In this work, LLMs are not used for conventional forecasting or statistical modeling of demand; rather, they are leveraged as controlled generative agents that produce scenario-aware synthetic representations of household energy consumption under explicitly defined cultural, temporal, and contextual assumptions. Inspired by \cite{Karpathy_2025}, we employ a multi-agent LLM Council in which distinct models operate in complementary roles under explicitly defined constraints. This improves consistency and contextual coverage by relying on an ensemble of models rather than a single model’s capabilities. Specifically, we aim to integrate cultural, seasonal, and behavioral contexts into household energy consumption patterns. We acknowledge that LLM-generated data faces inherent challenges in reproducing physically grounded dynamics. We therefore position \textit{WattCouncil} as a tool for generating controlled synthetic scenarios for analysis and benchmarking, and as a complement to physical simulators and to demand forecasting or optimization. Our contributions consist of:
(i) \textit{WattCouncil}, a multi-agent LLM-based framework for governed synthetic energy data generation;
(ii) a staged pipeline with explicit auditing and bounded regeneration;
(iii) an evaluation protocol based on similarity to real smart meter data;
(iv) ablation studies on the components of our data generation pipeline;
(v) \href{https://github.com/Singularity-AI-Lab/wattcouncil}{open-access} code of our pipeline with clear instructions on how to run it, hoping that it could be a contribution to future researchers who want to explore the area of LLM-based data synthesis.

The remainder of the paper is structured as follows. \autoref{Sec-Literature} reviews related work in household energy modeling, synthetic data generation, and the use of large language models in structured simulation contexts. \autoref{Sec-Methodology} introduces the \textit{WattCouncil} framework, describing the staged generation pipeline, constraint formulation, and governance mechanisms. \autoref{Sec-Results} presents a quantitative and qualitative evaluation of the generated consumption profiles, including comparisons with a real-world smart meter dataset, while \autoref{Sec-Ablation} analyzes the impact of key design choices through targeted ablation studies. \autoref{Sec-Limitations} discusses current limitations of the proposed approach, and \autoref{Sec-Conclusion} concludes the paper with a summary of findings and directions for future work.
\section{Related Work}\label{Sec-Literature}

Traditionally, researchers enabled access to relevant data by conducting in-place studies for a defined period and gathering as much data as possible from real-world settings using sensors. Despite sustained efforts to collect and release residential electricity datasets, the resulting landscape remains highly fragmented across geography, duration, resolution, and household coverage. As pointed out in \cite{cruz2024pattern}, which found that in general, datasets varied considerably, especially in countries such as the USA, the UK, and India, and exhibited several sources of bias. Widely used datasets such as REDD \cite{kolter2011redd}, UK-DALE \cite{kelly2015uk}, REFIT \cite{murray2017electrical}, AMPds \cite{makonin2013ampds}, BLUED \cite{filip2011blued}, iAWE \cite{batra2013s}, and ECO \cite{beckel2014eco} provide high-resolution appliance-level measurements but are typically limited to a small number of households and short observation periods, thereby limiting their representativeness. On the other hand, larger-scale datasets such as Dataport (Pecan Street) \cite{parson2015dataport}, RECS \cite{EIA2020RECS}, and CLNRPD \cite{clnr2015closedown} offer broader population coverage but often lack fine-grained temporal resolution, appliance-level detail, or continuous monitoring across contexts. 

Household energy research has documented that residential electricity consumption is influenced by a combination of demographic, behavioral, cultural, and socio-economic factors. Empirical studies identify household size, income, appliance ownership, and occupancy patterns as significant drivers of energy use variability \cite{jones2015determinants}. For instance, larger household size has consistently been associated with higher electricity use, while income and dwelling-related characteristics (such as the number of bedrooms and heating systems) account for substantial differences in consumption across households \cite{jones2015determinants}. Other research has highlighted that socio-economic and regional characteristics result in heterogeneous consumption patterns that vary between geographic contexts and policy environments. This suggests that household electricity demand is shaped by interacting social and structural determinants rather than physical building attributes alone \cite{kostakis2020socio, najeeb2024determinants}. The study in \cite{moraEnergyConsumptionResidential2018} highlights the significant impact of occupant behavior, socio-economic characteristics, and preferences (e.g., set-point temperatures and ventilation durations) on residential energy consumption. It emphasizes that variations in energy demand are influenced more by contextual factors, such as family composition and dwelling usage patterns, than by physical attributes, such as house size or floor area. 

In \cite{duApplianceCommitmentHousehold2011}, the authors present an algorithm that schedules household appliances based on consumption forecasts and user comfort settings, aiming to optimize objectives like minimizing costs or maximizing comfort. This underscores the importance of accurately modeling intra-family dynamics to simulate appliance usage patterns effectively. Building upon this, \cite{NICHOLLS2015116} examines how family interactions, such as shared activities and the presence of children, influence peak electricity demand times. The study highlights the need to incorporate family dynamics into energy consumption models to accurately predict and manage peak demand periods. Several studies have leveraged weather simulation frameworks to assess HVAC performance under various climatic scenarios. For example, \cite{xuEffectsCoolMaterials2024} analyzed the impact of urban heat islands on residential cooling demand, showing how microclimatic factors intensify energy consumption in densely populated regions. Despite this substantial body of empirical work, many modeling approaches traditionally used in energy systems have focused primarily on physical parameters or historical load profiles, with limited integration of detailed socio-demographic and behavioral insights into analytical or generative frameworks \cite{crosbieHOUSEHOLDENERGYSTUDIES2006}. This gap motivates the development of methods that can represent contextually grounded consumption variability when generating or analyzing residential energy data.

The usage of LLMs is transforming how domain-specific tasks are approached, particularly in generating human-like, context-aware outputs. In \cite{busterSupportingEnergyPolicy2024}, researchers from the National Renewable Energy Laboratory (NREL) introduced a framework combining LLMs with decision-tree logic to guarantee compliance with zoning policies for renewable energy siting. The reported accuracy suggests that such approaches can be useful in downstream modeling. In \cite{almashor2024can}, the authors explored whether LLM-based agents can synthesize household energy consumption, building upon the environment provided by \cite{park2023generative}. Although they obtained some reasonable load patterns, they were limited by the capabilities of the simulation they used. Their approach focused on parsing keywords in agent conversations, associating them with energy consumption, and then summing the household load. Along the same line of work, \cite{takrouri2025knowledge} introduced a multi-stage methodology to synthesize load profiles from different LLMs using a pipeline that considers family structure definition, weather identification, weather synthesis, and load pattern generation, all conditioned on a given geography and time period. Building upon that work, \cite{chetty2025llm} presents a methodology that enhances the process by integrating memory capacities and a rule-based validation before constructing the final load profile. These studies highlight the potential of LLMs to streamline data access and policy analysis in household energy modeling. 

However, significant gaps persist in integrating these components into a cohesive framework. Current approaches often rely on limited assumptions about family structures and behaviors, neglecting the nuanced and culturally specific dynamics necessary for accurate simulations. In particular, existing work rarely separates macro-level scenario specification (e.g., season/climate regime, calendar rules, household composition, and social context) from micro-level behavioral realization at the hourly scale (occupancy/activity and appliance-use traces). This limits explainability and controllability.
\section{Methodology}\label{Sec-Methodology}

\subsection{WattCouncil: An LLM-Based Framework For Governance}\label{Sec-Council}

The research community has increasingly supported the idea that coordinated outputs from many LLMs could outperform a single model. Works such as \cite{hu2025efficientdynamicensemblingmultiple} showed that this approach, known in the literature as model ensembling, outperformed individual models in reasoning tasks. More sophisticated works, such as \cite{gao2025strategic}, proposed a knowledge-transfer method to optimize model ensemble performance in sequential decision-making tasks. Finally, a recently published resource in \cite{Karpathy_2025} introduced the idea of LLM councils, which consist of several models (with different roles) that reach consensus on a given task. 

\begin{figure}[tb]
    \centering
    \includegraphics[width=0.95\columnwidth]{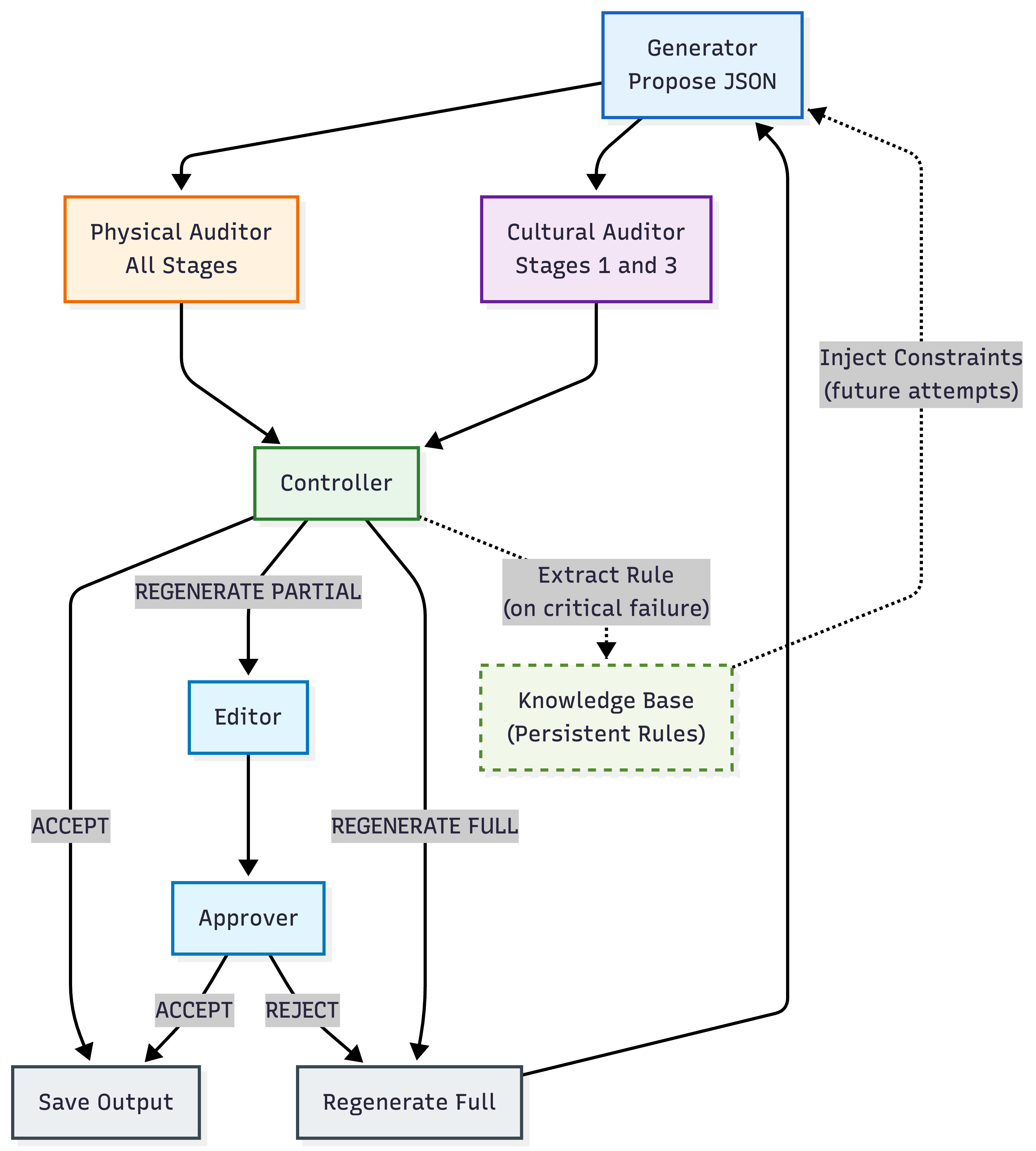}
    \caption{\textit{WattCouncil} governance and decision flow. A generator proposes schema-constrained JSON artifacts at each stage. Stage-specific auditors validate cultural plausibility (Stages 1 and 3) and physical consistency (all stages). A controller applies fixed decision rules to either accept, partially regenerate via constrained editing and approval, or fully regenerate.}
    \label{fig:council}
    \Description{A flow diagram describing the governance and decision flow of our framework called WattCouncil.}
\end{figure}

Inspired by the aforementioned works, we introduce \textit{WattCouncil}, a \emph{governed LLM Council} that employs role-specialized agents to generate structured synthetic household energy data under explicit constraints. Rather than relying on a single language model to produce end-to-end outputs, the council decomposes generation into a sequence of staged decisions, each subject to a data schema validation, auditing, and explicit acceptance criteria. In contrast to autonomous conversational use, the LLMs operate as constrained generators inside a pipeline whose control flow, stopping conditions, and artifact interfaces are fully specified. \autoref{fig:council} summarizes the governance logic and decision flow enforced at every level.

\begin{figure}[tb]
    \centering
    \includegraphics[width=0.95\columnwidth]{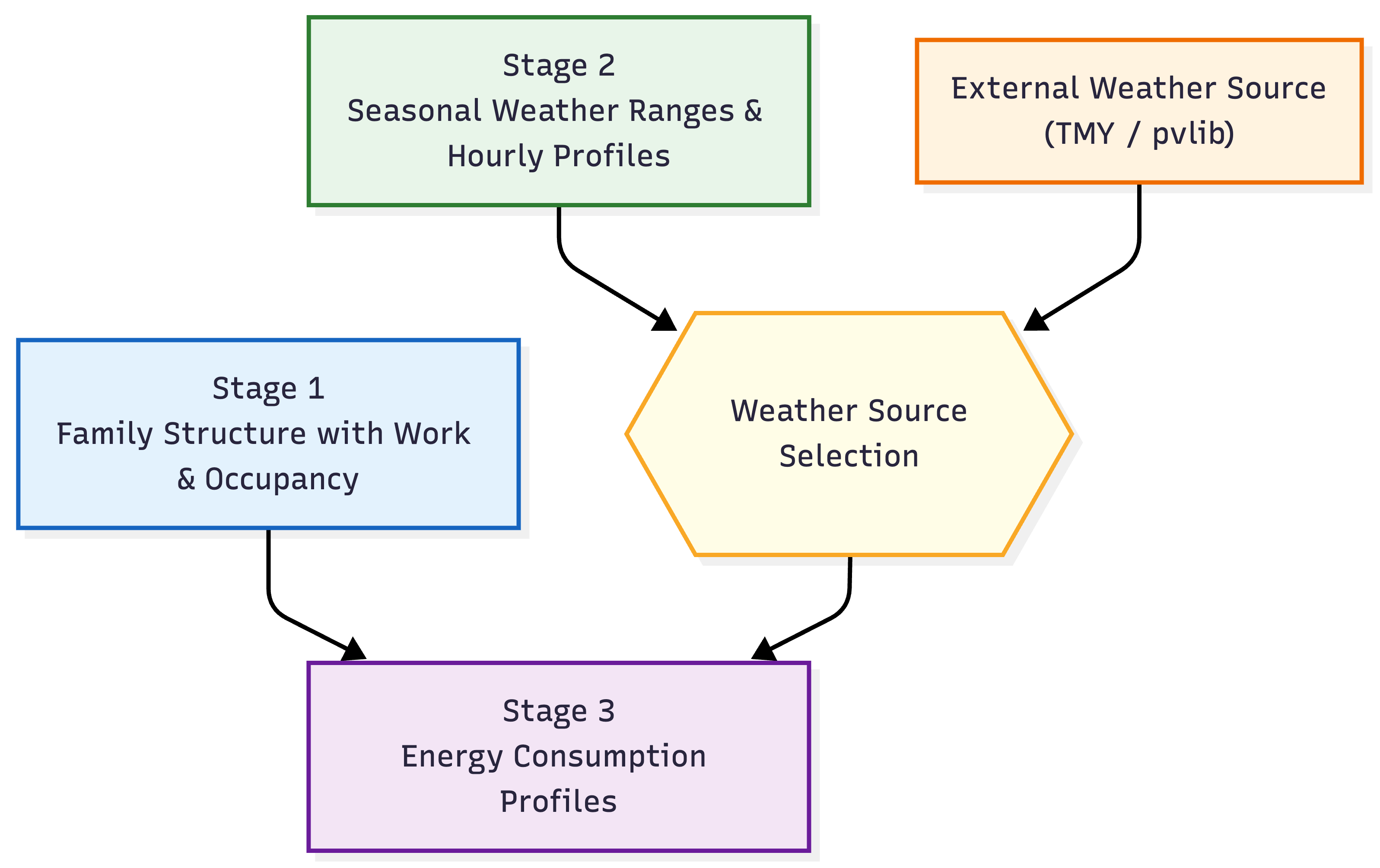}
    \caption{\textbf{Three-stage household energy generation pipeline with modular weather sourcing.} Household energy profiles are generated through three stages: (1) household structure and occupancy assignment, (2) weather profile generation or ingestion, and (3) energy consumption synthesis. Weather inputs may be produced by an LLM-based generator or replaced with externally sourced Typical Meteorological Year data; both paths converge at Stage~3 to enable controlled ablation between generative and hybrid settings. All stages operate under the governance of the LLM Council (\autoref{fig:council}).}
    \Description{A flow diagram of a three-stage household energy generation pipeline. Stage 1 generates household structure and occupancy. Stage 2 provides weather inputs, either from an internal generative process or an external Typical Meteorological Year source. Both weather paths merge at Stage 3, where household energy consumption profiles are produced.}
    \label{fig:pipeline}
\end{figure}

The council is implemented as a three-stage pipeline with explicit artifact interfaces as illustrated in \autoref{fig:pipeline}. The council comprises three components: (i) household structure generation, with assignment of work and occupancy regimes; (ii) definition of seasonal weather ranges, with hourly weather instantiation; (iii) household energy consumption modeling. Each stage consumes only outputs that are \emph{validated} in the operational sense, in other words, outputs that generated a schema-compliant JSON that has been audited and accepted by the controller/approver. Individual stages can be selectively enabled, disabled, or replaced by external data sources without affecting the integrity of the remaining pipeline.

The council architecture assigns fixed, non-overlapping roles to multiple LLMs, enforcing strict task separation. A primary \emph{generator} proposes candidate structured outputs at each stage of the pipeline, which are evaluated by designated role-based auditors depending on the stage context. For stages involving social structure, behavior, or narrative coherence (Stages 1 and 3), a \emph{cultural auditor} assesses alignment with country-specific social norms, and a \emph{physical auditor} verifies numerical consistency and adherence to physical or environmental constraints across all stages. For Stage 2, only the \emph{physical auditor} is applied, as cultural evaluation is not meaningful in contexts such as seasonal weather-range definition and hourly weather instantiation. Auditors return structured JSON reports with a severity score, following a shared scale (\texttt{LOW}/\texttt{MEDIUM}/\texttt{HIGH}) that indicates the level of compliance with the conditions for each stage. We maintain a fixed generation logic to achieve reproducibility by pinning model versions, prompts, and schemas, and logging all intermediate artifacts, while fixing inference settings (e.g., temperature and sampling parameters). 

Finally, a \emph{Council Controller} acts as the sole decision-making authority in the council. Based on the generator output and the relevant audit results, it selects one of three actions: accepting the results of the pipeline execution (\texttt{ACCEPT}), triggering a targeted correction phase handled by an \emph{editor} followed by independent verification by an \emph{approver} (\texttt{REGENERATE PARTIAL}), or discarding the execution and restart the process from scratch (\texttt{REGENERATE FULL}). The framework supports a scoped \emph{rule memory} that records recurring failure patterns (e.g., repeated schema violations or constraint failures) and the corresponding corrective instructions. When enabled, this memory is treated as a versioned artifact: any rule additions are logged so that subsequent generations can be replayed under the same rule set.

The \textit{WattCouncil} framework is intentionally model-agnostic, assigning specific language models to each council role based on functional suitability (e.g., cultural reasoning, numerical consistency, or structured editing), while keeping these assignments fixed across all experiments. The framework does not benchmark or compare individual language models. Instead, evaluation is performed at the system level, where behavior emerges from role specialization, constraint enforcement, and deterministic governance rather than from any single model’s capabilities. \autoref{tab:council_config} summarizes the models used in this study.

\begin{table*}[tb]
\centering
\caption{LLM Council role configuration and approximate model scale. The hyperparameter $\tau$ corresponds to the temperature of the softmax function in the output of the models. Higher temperatures imply greater randomness in generation, encouraging diversity, while lower temperatures yield more consistent outputs.}
\label{tab:council_config}

\begin{tabular}{l l l l l l}
    \toprule
    \textbf{Council Role} & \textbf{Model} & $\boldsymbol{\tau}$ & \textbf{Max. Tokens} & \textbf{Primary Function} \\
    \midrule
    Generator &
    Gemini 2.5 Flash &
    0.7 & 16000 &
    Primary structured generation \\
    
    Cultural Auditor &
    Llama-4 Maverick 17B &
    0.3 & 4000 &
    Cultural and social plausibility \\
    
    Physical Auditor &
    Claude 4 Sonnet &
    0.3 & 4000 &
    Numerical and physical validation \\
    
    Editor &
    Llama 3.3 70B &
    0.4 & 16000 &
    Targeted partial regeneration \\
    
    Approver &
    Mistral Small 3.2 &
    0.3 & 4000 &
    Editor's output verification \\
    
    Controller&
    Qwen2.5-72B-Instruct &
    0.4 & 4000 &
    Decision-making \\
    
    \bottomrule
\end{tabular}
\end{table*}

In LLMs, the \emph{temperature} hyperparameter ($\tau$) controls sampling randomness. Intuitively, temperature determines how willing the model is to explore alternative plausible outputs rather than repeatedly selecting the most likely continuation: lower values make the model more conservative and predictable, while higher values allow greater variation. Formally, given a vector of logits $\mathbf{z} = (z_1, \dots, z_K)$ over a vocabulary of size $K$, the probability of sampling token $i$ is computed using a temperature-scaled softmax,
\begin{equation}
p_i = \frac{\exp\left(z_i / \tau\right)}{\sum_{j=1}^{K} \exp\left(z_j / \tau\right)},
\end{equation}
where lower $\tau$ concentrates probability on high-scoring tokens and higher $\tau$ increases output diversity. In \textit{WattCouncil}, temperature controls local generation variability, while global consistency is enforced through auditing, bounded regeneration, and explicit acceptance criteria.

The temperature settings assigned to each council role reflect their functional responsibilities within the generation and governance pipeline. In particular, the \emph{Generator} is configured with a higher temperature than the other roles to encourage controlled creativity during the initial proposal of structured artifacts. At this stage, the objective is not strict verification but the exploration of plausible household structures, behaviors, and contextual configurations that are consistent with high-level constraints. A moderately elevated temperature increases output diversity, allowing the Generator to surface alternative yet reasonable scenarios that might not emerge under near-deterministic sampling. In contrast, downstream roles such as the Cultural Auditor, Physical Auditor, Editor, and Approver are assigned lower temperatures to prioritize consistency, precision, and adherence to constraints during evaluation and correction. This asymmetric temperature allocation reflects a deliberate separation of concerns: creative variability is permitted at the proposal stage, while rigor and stability are enforced during auditing and decision-making. Within the \textit{WattCouncil} framework, this design allows limited stochastic exploration without compromising overall governance, reproducibility, or physical and contextual validity.

\subsection{Integration of Typical Meteorological Weather (TMY)}

As stated before, the stages of \textit{WattCouncil}'s pipeline (\autoref{fig:pipeline}) are explicitly designed to be modular and can be replaced by externally sourced data without affecting the rest of the framework. This flexibility is particularly important for weather information, which is inherently complex and difficult to generate reliably from scratch due to its strong physical constraints, temporal structure, and geographic dependence. To address this, we investigate an alternative setup in which LLM-generated weather is replaced with a simple, safe baseline based on Typical Meteorological Year (TMY) data. 

A \emph{Typical Meteorological Year} is a synthesized weather dataset that represents statistically typical meteorological conditions for a specific geographic location over the course of a single year. Rather than reflecting a specific calendar year’s weather, a TMY is constructed by selecting the most representative months from a long historical record so that key variables such as temperature, solar radiation, humidity, and other meteorological indicators closely match long-term averages. TMY datasets are designed to capture typical patterns rather than extreme events (heat waves, cold snaps, or compound weather extremes); they are widely used in building energy simulation, renewable resource assessment, and engineering analyses as a representative baseline.

Although TMY data do not perfectly replicate year-to-year weather variability, they provide physically grounded, historically derived weather profiles that work as a reference for correctness. We implement this substitution using the Python library pvlib \cite{anderson2023pvlib}, which enables the generation of a full year of hourly weather variables for a given geographic location specified by latitude and longitude. This replacement decouples behavioral and energy modeling from weather synthesis, providing a physically grounded baseline for comparison. In the present study, weather conditioning focuses on representative seasonal patterns rather than explicit extreme-event stress testing.
\section{Experimental Results}\label{Sec-Results}

We evaluated the \textit{WattCouncil} framework as a \emph{scenario-oriented synthetic data generator}, assessing whether the generated household energy profiles reproduce realistic temporal structure and aggregate behavior when compared against real-world reference data, and whether key pipeline components, with a focus on weather data, can be substituted without substantially degrading downstream output.

\subsection{Reference Real-World Dataset}
The CER Electricity Data (Revised March 2012) \cite{cer_dataset_2024} is a rich, open dataset derived from the Commission for Energy Regulation's Smart Metering Project, a large-scale trial of electricity smart meters in Ireland. It comprises half-hourly electricity consumption measurements collected over more than 500 days (spanning from July 14, 2009, to December 31, 2010), capturing detailed load profiles for 4232 residential customers and 529 small-to-medium enterprises (which we do not consider in our study). In addition to the raw consumption time series, the trial includes extensive pre- and post-trial survey data on socio-demographic characteristics, household properties, and energy-use behaviors, enabling analysis that links load profiles to occupant and dwelling features. We selected this dataset as a reference because it reflects real consumer behavior under the determinants we introduced in our pipeline, enabling a direct comparison between synthetic and real data.

\subsubsection{Exploratory Data Analysis (EDA) of the CER Dataset.} Inspired by prior work \cite{guo2022predicting} that uses the CER dataset and combines the smart meter time series with socio-demographic metadata to study household energy behavior, we follow the same principle of jointly using consumption traces and survey-derived contextual attributes. We adopt the perspective that load profiles alone are insufficient to characterize household behavior without accounting for family composition, occupancy patterns, and socio-economic factors, all of which are explicitly modeled in our framework. While that study used K-means (an unsupervised clustering method) to identify latent customer groups based on load shape similarity, we intentionally opted for a different approach that explicitly groups clients based on responses to selected questions in the surveys.

\begin{table}[tb]
    \centering
    \caption{Selected residential customer groups from the CER dataset, stratified by household size, house type, and household composition. House types are encoded as: (1) Apartment, (2) Semi-detached, (3) Detached, (4) Terraced, and (5) Bungalow. Household composition categories are: (1) Single occupant, (2) Adults only (all members over 15 years), and (3) Adults with children under 15 years. Apartment households did not appear among the top 20 most frequent groups and were therefore excluded.}
    \label{tab:cer_customer_groups}
        \begin{tabular}{c c c c c c}
            \toprule
            \textbf{Group} & \makecell{\textbf{Household} \\ \textbf{Size}} & 
            \makecell{\textbf{House} \\ \textbf{Type}} & \makecell{\textbf{Household} \\ \textbf{Type}} & \makecell{\textbf{Group} \\ \textbf{Size}} \\
            \midrule
            1 & 1 person & 5 & 1 & 171 \\
            2 & 2 people & 4 & 2 & 173 \\
            3 & 3 people & 2 & 3 & 80 \\
            4 & 3 people & 3 & 2 & 90 \\
            5 & 4 people & 3 & 3 & 113 \\
            \bottomrule
        \end{tabular}
\end{table}

Following the design and operational principles of \textit{WattCouncil}, the cluster membership carries limited semantic meaning for conditioned generation; in other words, including an unlabeled cluster in the context does not provide interpretable guidance for scenario synthesis. Instead, we define explicit household groups based on socio-economic and survey attributes available in the CER dataset, selecting variables that can be directly used as conditioning inputs within our pipeline, as summarized in \autoref{tab:cer_customer_groups}. From these groups, we identify representative subsets of households and extract their aggregated characteristics to ground the generation process. This design choice ensures that insights derived during the EDA stage explicitly constrain the generative mechanism, enabling systematic and interpretable comparisons between real consumption data and LLM-generated profiles produced by our pipeline.

\subsection{Comparison Between Synthetic and Real-World Energy Consumption Data}\label{Sec-Synth-vs-Real-Comparison}

We evaluate the realism of the synthetic household energy consumption profiles generated by \textit{WattCouncil} by comparing them against the CER dataset. The objective is to assess whether the proposed pipeline reproduces residential consumption patterns observed in real data across daily and seasonal time scales.

Based on the selected residential groups reported in \autoref{tab:cer_customer_groups}, we condition the generation process on these 5 representative groups. The pipeline generates multiple synthetic households that match the groups’ characteristics, and daily consumption profiles are produced under multiple weather conditions across all meteorological seasons. These synthetic profiles are then compared against the corresponding subset of real households from the CER dataset to assess behavioral and temporal consistency.

\subsubsection{Seasonal Profile Construction.}

For the synthetic data, we aggregated energy consumption across all family-weather combinations within each season. We compute hourly means, standard deviations, and sample counts, yielding a representative 24-hour profile for each season. For real data, we define seasons by calendar months in the Northern Hemisphere: winter (December–February), spring (March–May), summer (June–August), and autumn (September–November). We grouped hourly consumption values by season and hour of day across all selected customers and computed the corresponding summary statistics.

\subsubsection{Visualization and Similarity Metrics.}

\begin{figure*}[tb]
    \centering
    \includegraphics[width=\textwidth]{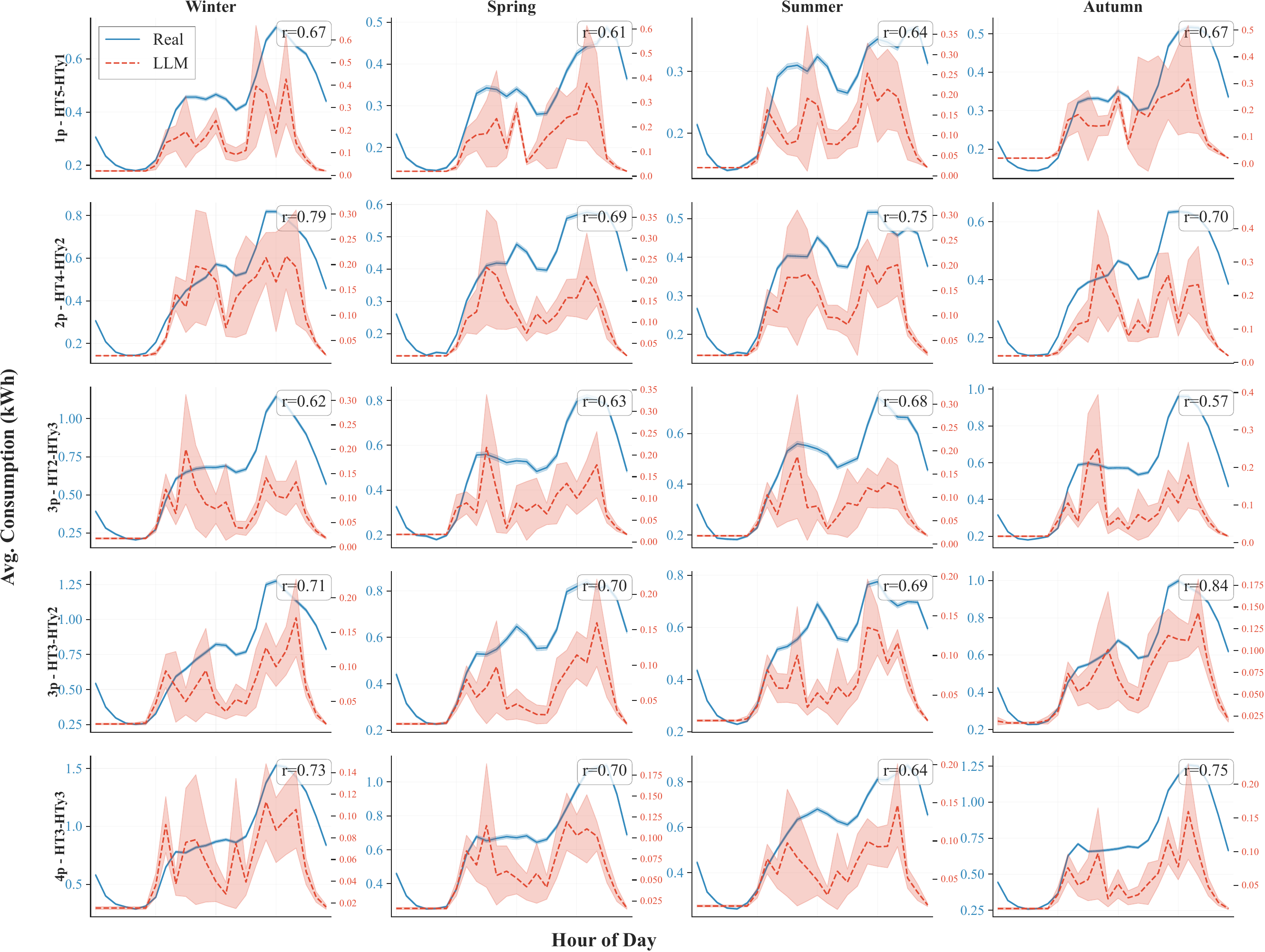}
    \caption{Comparison between CER dataset and \textit{WattCouncil} seasonal daily profiles. We generate daily profiles using available data from the 5 groups in \autoref{tab:cer_customer_groups}, totaling 627 customers. We include the Pearson correlation coefficient ($r$) to quantify the similarity of daily load shapes.}
    \label{fig:llm_vs_real_comparison}
    \Description{Comparison between CER dataset and WattCouncil seasonal daily profiles with Pearson correlation coefficients to quantify the similarity of daily load shapes.}
\end{figure*}

We visualize seasonal comparisons using four horizontally arranged subplots as shown in \autoref{fig:llm_vs_real_comparison} corresponding to winter, spring, summer, and autumn. Each subplot displays the mean hourly consumption profiles for synthetic and real data, accompanied by shaded confidence intervals. For this \textit{WattCouncil} dataset, we generated electricity consumption profiles for five household groups by combining multiple controlled variations. Specifically, we selected 5 representative family groups and instantiated 10 employment and occupancy variants per group. Each configuration was evaluated across 4 seasons and 5 independent weather realizations per season, resulting in a total of $5 \times 10 \times 4 \times 5 = 1000$ synthetic household profiles. We note that the apparent smoothness difference in \autoref{fig:llm_vs_real_comparison} may arise from the original axis scaling and the differences in aggregation: the real profiles are averaged over a larger number of households, whereas the synthetic profiles are derived from a smaller set of generated household configurations.

In addition to visual inspection, \autoref{tab:consolidated_metrics} reports the quantitative error and similarity metrics for each season. We use Pearson's correlation to assess temporal shape similarity and peak timing alignment, whereas mean absolute error (MAE), root mean squared error (RMSE), and mean absolute percentage error (MAPE) quantify magnitude mismatch. This combination allowed separating whether the synthetic profiles reproduce realistic daily temporal structure, and whether they recover realistic absolute consumption levels. Results show pattern--level similarities but clear magnitude discrepancies, indicating that matching the consumption dimension depends on aspects we might not have considered in our pipeline. This distinction is important: realistic timing patterns can emerge from context-aware occupancy and activity modeling even when the final consumption scale remains imperfect.

\begin{table}[tbp]
\centering
\caption{Comparison metrics between synthetic LLM-generated and real customer energy consumption profiles across demographic groups and seasons.}
\label{tab:consolidated_metrics}
\begin{tabular}{clllll}
\toprule
\textbf{Group} & \textbf{Season} & \makecell{\textbf{MAE} \\ \textbf{(kWh)}} & \makecell{\textbf{RMSE} \\ \textbf{(kWh)}} & \makecell{\textbf{MAPE} \\ \textbf{(\%)}} & \textbf{Corr.} \\
\midrule
1 
 & Winter & 0.294 & 0.319 & 72.76 & 0.670 \\
 & Spring & 0.177 & 0.200 & 62.07 & 0.611 \\
 & Summer & 0.170 & 0.181 & 65.98 & 0.639 \\
 & Autumn & 0.193 & 0.213 & 63.97 & 0.670 \\
\midrule
2
 & Winter & 0.362 & 0.398 & 79.52 & 0.789 \\
 & Spring & 0.279 & 0.301 & 75.71 & 0.689 \\
 & Summer & 0.254 & 0.268 & 74.43 & 0.751 \\
 & Autumn & 0.277 & 0.301 & 74.22 & 0.699 \\
\midrule
3
 & Winter & 0.551 & 0.607 & 89.21 & 0.624 \\
 & Spring & 0.418 & 0.452 & 85.59 & 0.634 \\
 & Summer & 0.387 & 0.414 & 85.70 & 0.677 \\
 & Autumn & 0.459 & 0.506 & 86.66 & 0.569 \\
\midrule
4
 & Winter & 0.659 & 0.721 & 91.69 & 0.714 \\
 & Spring & 0.492 & 0.522 & 90.37 & 0.702 \\
 & Summer & 0.471 & 0.494 & 89.89 & 0.687 \\
 & Autumn & 0.525 & 0.567 & 90.08 & 0.838 \\
\midrule
5
 & Winter & 0.789 & 0.868 & 93.98 & 0.726 \\
 & Spring & 0.590 & 0.635 & 91.66 & 0.703 \\
 & Summer & 0.516 & 0.547 & 91.03 & 0.643 \\
 & Autumn & 0.644 & 0.704 & 93.01 & 0.749 \\
\bottomrule
\end{tabular}
\end{table}

\section{Ablation Studies on the Pipeline}\label{Sec-Ablation}

The ablation is evaluated at two levels: \autoref{Sec-Results-Weather} compares LLM- and TMY-derived weather profiles (seasonal statistics and daily patterns), and \autoref{Sec-Results-Energy} quantifies changes in downstream demand patterns deriving from changing the source of weather data.

\begin{figure*}[tb]
    \centering
    \includegraphics[width=0.92\textwidth]{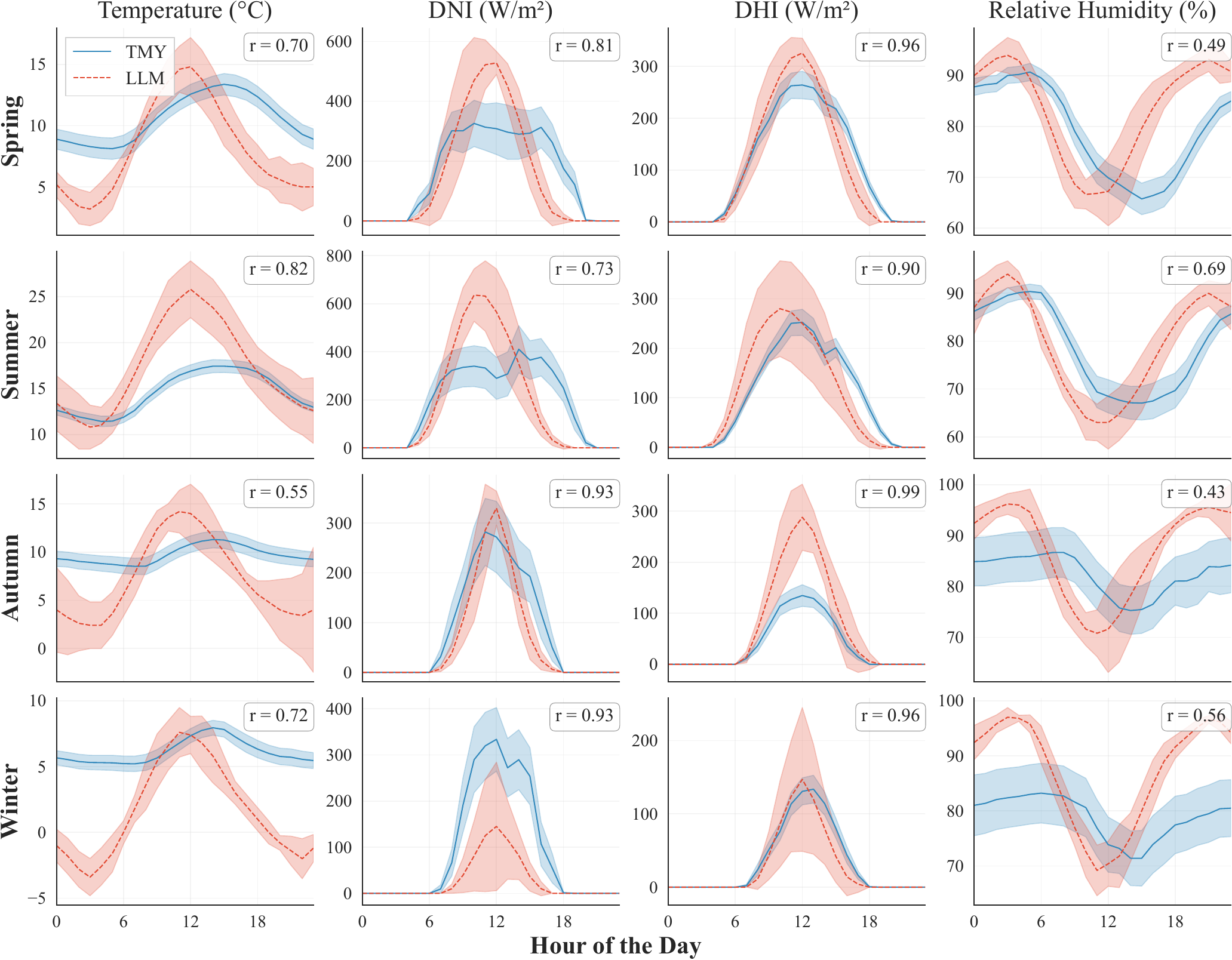}
    \caption{\textbf{LLM-based vs TMY weather comparison. This figure shows a visual comparison of the average daily per-season profiles for the environmental variables we consider: temperature, direct normal irradiance (DNI), diffuse horizontal irradiance (DHI), and relative humidity (RH). We measure the correlation between the average profiles to assess profile similarity.}
    \label{fig:weather_grid}}
    \Description{Grid view that compares LLM-generated versus TMY temperature, direct normal irradiance, diffuse horizontal irradiance, and relative humidity.}
\end{figure*}

\subsection{Weather Profile Comparison}\label{Sec-Results-Weather}

We first assess whether the proposed pipeline produces physically and temporally plausible weather inputs by comparing LLM-generated hourly weather profiles with externally sourced TMY data for Ireland, which matches the geographic scope and reference year (2009) of the real-world dataset used in our study. This analysis isolates the behavior of the weather-generation component (Stage 2) from household structure and behavioral modeling.

\autoref{fig:weather_grid} visually compares the average daily profiles of environmental variables generated with the LLM against the TMY reference for Dublin, Ireland, which we assume is uniform across the territory (the dataset does not provide household locations to preserve privacy). We used 5 LLM-generated profiles and 60 TMY profiles (which are faster to sample) to build 95\% confidence intervals using the t-distribution. Additionally, we computed Pearson correlation coefficients ($r$) between LLM-generated and TMY weather profiles across all seasons and meteorological variables. The results suggest that LLM-generated data align with the TMY reference, with correlation increasing in spring and summer and decreasing in autumn and winter. 

\begin{table}[tb]
    \centering
    \caption{Pearson correlation coefficient between LLM-generated and TMY profiles.}
    \label{tab:corr_llm_vs_tmy_weather}
    \begin{tabular}{l l l l l l}
        \toprule
        \textbf{Season} & \textbf{Temp.} & \textbf{DNI} & \textbf{DHI} & $\textbf{RH}$ \\
        \midrule
        Winter & 0.724 & 0.933 & 0.961 & 0.563 \\
        Spring & 0.700 & 0.810 & 0.961 & 0.491 \\
        Summer & 0.822 & 0.728 & 0.903 & 0.694 \\
        Autumn & 0.547 & 0.930 & 0.994 & 0.428 \\
        \bottomrule
    \end{tabular}
\end{table}

We summarize the correlation results in \autoref{tab:corr_llm_vs_tmy_weather}. These results indicate that directly generating weather profiles with general-purpose LLMs can introduce avoidable inaccuracies, primarily because of the physical complexity and high temporal precision required for meteorological modeling. In this setting, LLMs are less well-suited than physics-informed or data-driven weather models, rather than being inherently incapable. 

Importantly, this limitation does not undermine the \textit{WattCouncil} framework; instead, it motivates the use of reliable external weather sources, such as TMY data, to condition consumption and generation. Moreover, specialized climate models—such as NVIDIA's Earth-2 platform \cite{Earth2Studio_Contributors_NVIDIA_Earth2Studio_2024}—represent promising avenues for integrating richer, physically grounded weather context into the pipeline without compromising consistency or realism.

\begin{figure*}[tb]
    \centering
    \includegraphics[width=0.95\textwidth]{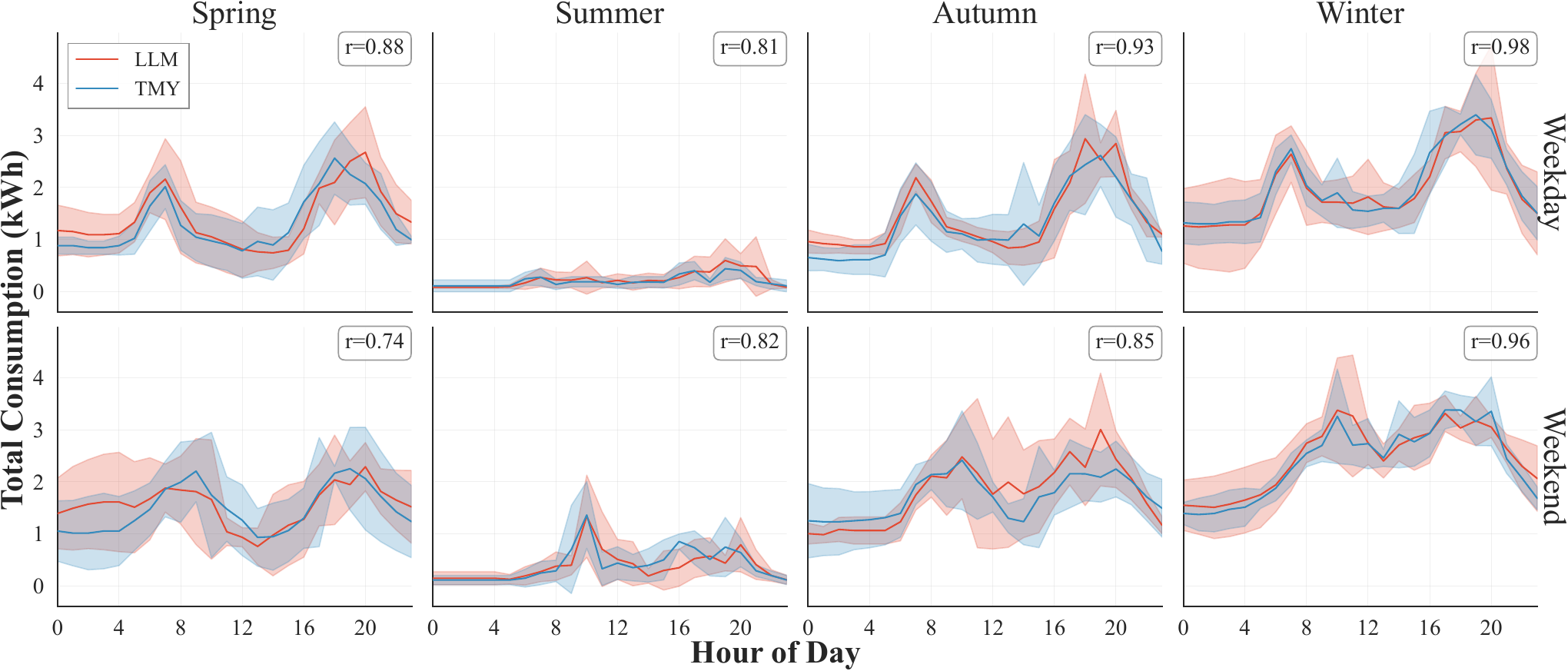}
    \caption{
    Impact of weather sourcing on household energy consumption. Comparison of hourly electricity demand profiles generated using LLM-based weather and externally sourced TMY weather for a fixed Irish household with adults and children under 15 years. Profiles are shown for weekdays and weekends across all four seasons, with shaded bands indicating variability across five realizations. Pearson correlation coefficients ($r$) quantify the similarity between profiles under the two weather sources.}
    \label{fig:fix_family_weather}
    \Description{Grid of hourly electricity consumption profiles comparing LLM-generated weather and TMY weather inputs for a fixed Irish household with adults and children under 15 years. Columns correspond to seasons (spring, summer, autumn, winter), while rows distinguish weekday and weekend profiles. The curves exhibit closely aligned load shapes and peak timing, with Pearson correlation coefficients ranging from approximately 0.74 to 0.98, indicating strong agreement between the two weather sourcing strategies.}
\end{figure*}

\subsection{Impact of Weather Sourcing on Energy Consumption}
\label{Sec-Results-Energy}

To isolate the effect of weather sourcing, all non-meteorological inputs are held constant across experiments. \autoref{tab:fixed_household_ablation} summarizes the household configuration used in this study, including composition, occupancy regime, dwelling type, and appliance set.

Electricity consumption profiles are generated under two settings: LLM-generated weather inputs and externally sourced TMY weather. \autoref{fig:fix_family_weather} compares weekday and weekend demand profiles across all four seasons for five realizations of the same household configuration. The results exhibit closely aligned load shapes and peak timing, indicating that downstream consumption dynamics remain stable under changes in weather sourcing. This observation is supported by high Pearson correlation coefficients across seasons. Differences between the two settings primarily appear in the width of uncertainty bands rather than systematic shifts in mean demand.

Overall, these results suggest that the proposed pipeline is robust to the choice of weather source at the household energy-demand level. LLM-generated weather preserves realistic demand structure and variability, while enabling scenario diversity beyond what is attainable from a single deterministic reference year.

One possible explanation for the muted summer demand is the combined effect of governance constraints, where cooling and heating are repeatedly disabled during regeneration, and behavioral assumptions associated with the fixed household configuration, leading to intentionally conservative summer usage patterns.

Notably, across repeated runs, we observe that summer scenarios consistently trigger additional regeneration and editing steps within the council, resulting in nearly twice the generation time compared to other seasons, further reflecting the stricter constraint enforcement under summer conditions.

\begin{table}[tb]
\centering
\caption{Fixed household configuration used in the weather-sourcing ablation study. All attributes are held constant while varying the source of weather inputs.}
\label{tab:fixed_household_ablation}

\begin{tabularx}{\linewidth}{%
    >{\raggedright\arraybackslash}p{0.38\linewidth}
    >{\raggedright\arraybackslash}X}
\toprule
\textbf{Attribute} & \textbf{Value} \\
\midrule
Country & Ireland \\
Household type & Adults with children $<$ 15 years \\
House type & Detached house \\
Number of occupants & 4 \\
Household composition & Father (office), Mother (hybrid), Son (student), Daughter (student) \\
Household work regime & Mixed \\
Weekday daytime occupancy & Partial \\
Chief income earner status & Employee \\
Appliances & Washing machine, Dishwasher, Oven, Refrigerator, Television, Microwave \\
\bottomrule
\end{tabularx}
\end{table}

\subsection{End-to-End Pipeline Execution Example}
\label{Sec-Results-Example}
To illustrate the behavior of the proposed framework in practice, we report an end-to-end execution trace for a single synthetic household processed through all pipeline stages. The household is evaluated under a single weather condition across the four seasons, with each season instantiated under both weekday and weekend assumptions. This results in 8 independent Stage 3 consumption-generation runs for the same household. During execution, the framework applies the LLM Council governance described in \autoref{fig:council} across all pipeline stages. When violations are detected at any stage (e.g., implausible household routines, inconsistent weather patterns, or HVAC heating active during summer conditions), the controller enforces regeneration with explicit, stage-specific corrective guidance. Because all detected violations triggered full regeneration rather than constrained editing, neither the Editor nor the Approver roles were activated in this execution. In the illustrated example, the first summer weekday attempt is rejected due to physically implausible heating usage and is subsequently regenerated and accepted after correction. This demonstrates how the framework actively enforces domain constraints rather than passively accepting generative outputs.

\subsubsection{Pipeline runtime analysis.}

\begin{table}[tb]
\centering
\caption{LLM Council execution statistics for a single-household end-to-end pipeline run.
The household is evaluated across four seasons with weekday and weekend variants, producing eight Stage~3 consumption outputs.}
\label{tab:pipeline-stats}
\begin{tabular}{lrrrr}
\toprule
\textbf{Council Role} & \textbf{Calls} & \textbf{Time (s)} & \textbf{In Tokens} & \textbf{Out Tokens} \\
\midrule
Generator        & 14 & 481.7 & 30{,}397 & 44{,}070 \\
Cultural Auditor & 10 & 15.1  & 43{,}581 & 309 \\
Physical Auditor & 14 & 28.7  & 56{,}808 & 549 \\
Controller       & 14 & 31.3  & 59{,}324 & 847 \\
Editor           & 0  & 0.0   & 0        & 0 \\
Approver         & 0  & 0.0   & 0        & 0 \\
\midrule
\textbf{Total}     & \textbf{52} & \textbf{556.8} & \textbf{190{,}110} & \textbf{54{,}775} \\
\bottomrule
\end{tabular}
\end{table}

\autoref{tab:pipeline-stats} summarizes the execution statistics for this illustrative run. Although only a single household is processed, the example reflects the governance overhead incurred during realistic operation: generating one household across four seasons and weekday/weekend variants required 52 model calls and 556.8 seconds of execution time, spanning generation, auditing, and decision-making steps. Most of this overhead is concentrated in the Generator and Physical Auditor roles, which dominate the synthesis and validation of structured hourly consumption profiles under explicit constraints, while cultural auditing and controller decisions contribute comparatively little. We do not report monetary costs, as they depend on the specific models, providers, and pricing schemes used at runtime, which may change over time; instead, we report token usage to enable reproducible cost estimation across different deployment settings. Representative execution traces for Stages~1--3 are shown next as annotated pipeline logs.

\noindent
\begin{center}
\includegraphics[width=0.87\columnwidth]{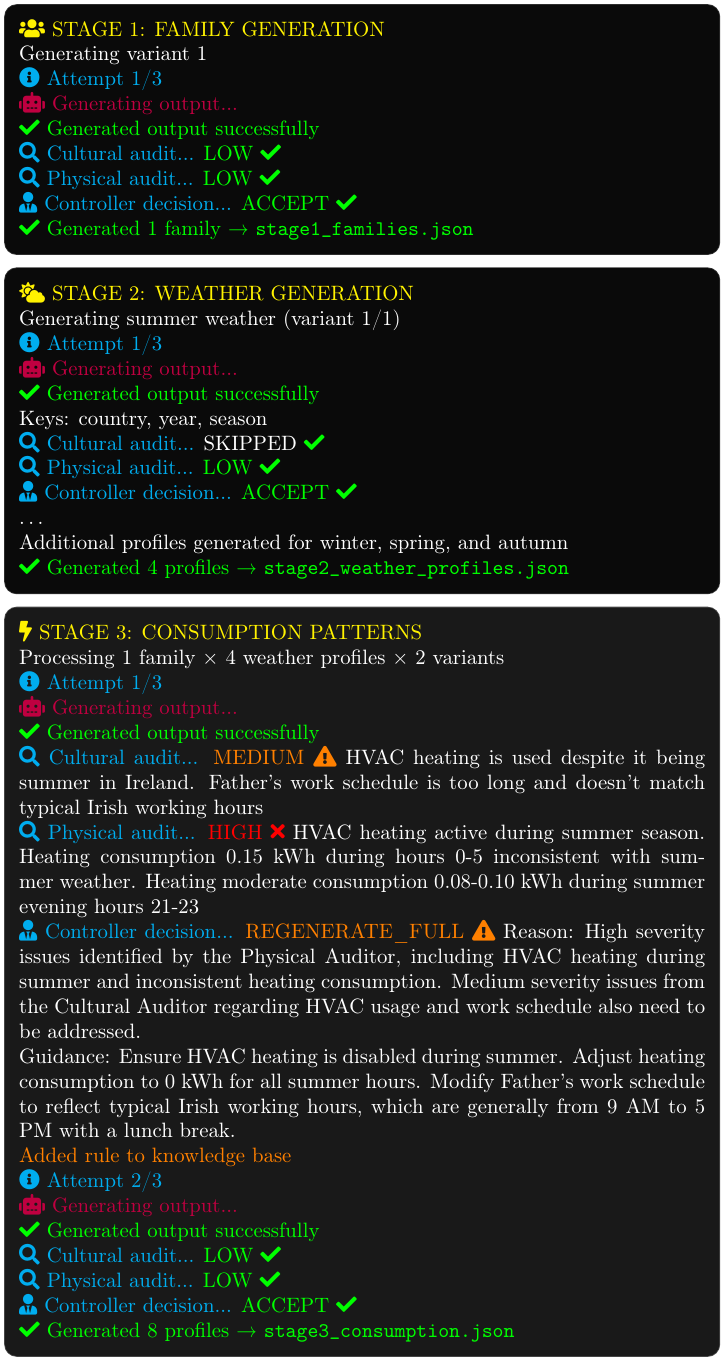}
\end{center}
\section{Limitations}\label{Sec-Limitations}

While the proposed framework aligns fairly with real-world residential electricity consumption patterns, several limitations should be noted.
\begin{enumerate}
    \item The empirical evaluation is limited to Irish households, given that the CER dataset was the only comprehensive study to provide sufficient socio-demographic metadata to create a systematic comparison. Rigorous validation against real data requires comparable datasets, constraining cross-country evaluation in this study.
    \item Comparisons against real data rely on aggregated seasonal and hourly statistics, which smooth household-level variability. Accordingly, the reported similarity metrics reflect population-level behavioral alignment rather than exact household-level replication.
    \item The weather ablation relies on a finite number of discrete realizations for both LLM-generated and TMY-derived inputs, which does not capture the full range of meteorological variability or extreme events.
    \item Governance mechanisms introduce additional computational overhead, reflecting a deliberate trade-off between controllability and generation speed.
    \item Our framework reduced a larger set of possible attributes available in the CER dataset metadata to three, aiming to keep our budget and computation tractable. The selection could introduce bias in how \textit{WattCouncil} generates load profiles.
    \item We assumed uniform weather in Ireland based on the low variance of weather stations, a similar approach to \cite{guo2022predicting}; this might be imprecise. Also, weather conditioning focuses on representative seasonal patterns rather than explicit extreme-event stress testing.
\end{enumerate}

\section{Conclusion}\label{Sec-Conclusion}

This paper introduced \textit{WattCouncil}, a governed, multi-agent LLM framework for generating structured synthetic household electricity consumption data under explicit cultural, temporal, and physical constraints. The proposed approach decomposes generation into a staged pipeline governed by role-specialized agents, explicit control logic, and governance via interfaces and control flows, with each stage producing auditable artifacts that meet explicit criteria. Evaluation against the CER smart meter dataset shows that the generated profiles reproduce the daily and seasonal demand structures fairly well and exhibit strong temporal alignment with real data across demographic groups. Ablation results further indicate that downstream consumption dynamics are robust to the choice of weather sourcing, with differences primarily affecting uncertainty rather than mean demand. While contextual and behavioral factors are central to reproducing temporal demand structure, the remaining magnitude gap indicates that physically grounded variables remain essential for realistic absolute consumption. A natural next step is therefore to incorporate stronger physical determinants, including dwelling-envelope properties, HVAC efficiency, appliance power ratings, and simplified thermal-response constraints. This would allow future versions of \textit{WattCouncil} to couple contextual plausibility with improved energy-scale realism. Future work could also extend the framework by incorporating a more diverse set of socio-economic conditioning attributes, evaluating additional regions, integrating physically informed climate models, and exploring tighter coupling with downstream optimization and planning tasks. The integration of explicit extreme-weather scenarios is also an important future direction, particularly in regions where heating and cooling demand are highly sensitive to rare meteorological events.
\clearpage
\bibliographystyle{ACM-Reference-Format}
\bibliography{ref}

\end{document}